\documentclass[pmlr,twocolumn,10pt]{jmlr} 

\usepackage{booktabs}
\usepackage{siunitx}

\usepackage[switch]{lineno}

\theorembodyfont{\upshape}
\theoremheaderfont{\scshape}
\theorempostheader{:}
\theoremsep{\newline}

\usepackage{makecell}

\usepackage{todonotes}

\title[Handling missing values in clinical machine learning: Insights from an expert study]{Handling missing values in clinical machine learning: \titlebreak Insights from an expert study}

\author{%
\Name{Lena Stempfle} \Email{stempfle@chalmers.se}\\
\addr Chalmers University of Technology \& Gothenburg University, Sweden
\AND
\Name{Arthur James} \Email{arthur.james@aphp.fr}\\
\addr Sorbonne University, GRC 29, AP-HP, DMU DREAM, \\Department of Anaesthesiology and Critical Care, Pitié-Salpêtrière Hospital, Paris, France
\AND
\Name{Julie Josse} \Email{julie.josse@inria.fr}\\
\addr PreMeDICaL Inria-Inserm, France
\AND
\Name{Tobias Gauss} \Email{tgauss@protonmail.com}\\
\addr Déchocage - Bloc des urgences, Pole Anesthésie-Réanimation, CHU Grenoble Alpes, France
\AND
\Name{Fredrik D. Johansson} \Email{fredrik.johansson@chalmers.se}\\
\addr Chalmers University of Technology \& Gothenburg University, Sweden
}

\begin{document}

\maketitle

\begin{abstract}
Inherently interpretable machine learning (IML) models offer valuable support for clinical decision-making but face challenges when features contain missing values. Traditional approaches, such as imputation or discarding incomplete records, are often impractical in scenarios where data is missing at test time. We surveyed 55 clinicians from 29 French trauma centers, collecting 20 complete responses to study their interaction with three IML models in a real-world clinical setting for predicting hemorrhagic shock with missing values. Our findings reveal that while clinicians recognize the value of interpretability and are familiar with common IML approaches, traditional imputation techniques often conflict with their intuition. Instead of imputing unobserved values, they rely on observed features combined with medical intuition and experience. As a result, methods that natively handle missing values are preferred. These findings underscore the need to integrate clinical reasoning into future IML models to enhance human-computer interaction.
\end{abstract}

\section{Introduction} \label{sec:intro} 
Missing values in healthcare data pose substantial challenges in predictive modeling, requiring effective methods to address them~\citep{austin2021missing, Wells2013}. These missing values occur in incomplete medical records and often arise intentionally, due to specific testing protocols, or unintentionally, due to time constraints or data collection errors~\citep{salgado2016missing}. The resulting gaps can significantly impact the reliability and applicability of predictive models, particularly at "test time" when they are used for decision-making.
Interpretable machine learning (IML) models have become essential in high-stakes decision-making areas such as healthcare~\citep{JMLR:v20:18-615}, sustainability~\citep{jin2022gein}, and criminal justice~\citep{wang2023pursuit, zhang2022interpretable}, where transparency and accountability are critical. These models offer insights into decision processes, ensuring that predictions can be understood and trusted by domain experts.
IMLs, such as decision trees, linear models, and rule-based models, are extensively studied for their intuitive structure and ease of interpretation~\citep{molnar2020interpretable, rudin2019stop}. They are widely used in tasks like classification, regression, and risk scoring~\citep{furnkranz2012foundations, margot2021new}.
As with most ML models, IML models to date rely on complete inputs, making them particularly vulnerable to missing data. In retrospective studies, a typical solution is complete-case analysis, where cases with missing values are excluded~\citep{little2019statistical}. While effective for analysis, this approach is unsuitable at test time, as it can leave many patients without predictions. An alternative is the \textit{impute-then-regress} strategy which imputes missing values before prediction, ensuring consistency between training and deployment tasks~\citep{rubin1976inference}. However, impute-then-regress is often biased when the missingness of predictive features depends on unobserved variables. 
Moreover, complex imputation methods can reduce interpretability, complicating deployment and decision-making. Despite extensive research on handling missing values~\citep{van2011mice, lemorvan20a_linear, chen2023missing, austin2021missing, luo2022evaluating, stempfle2024minty}, little is known about how medical domain experts interact with missing values when using prediction models.

\begin{figure}[t]
 \floatconts
 {fig:objectives}
 {\caption{Overview of study objectives, key survey question types, and findings on best practices for clinical interpretable machine learning with missing values at test time.}}
 {\includegraphics[width=1.\columnwidth]{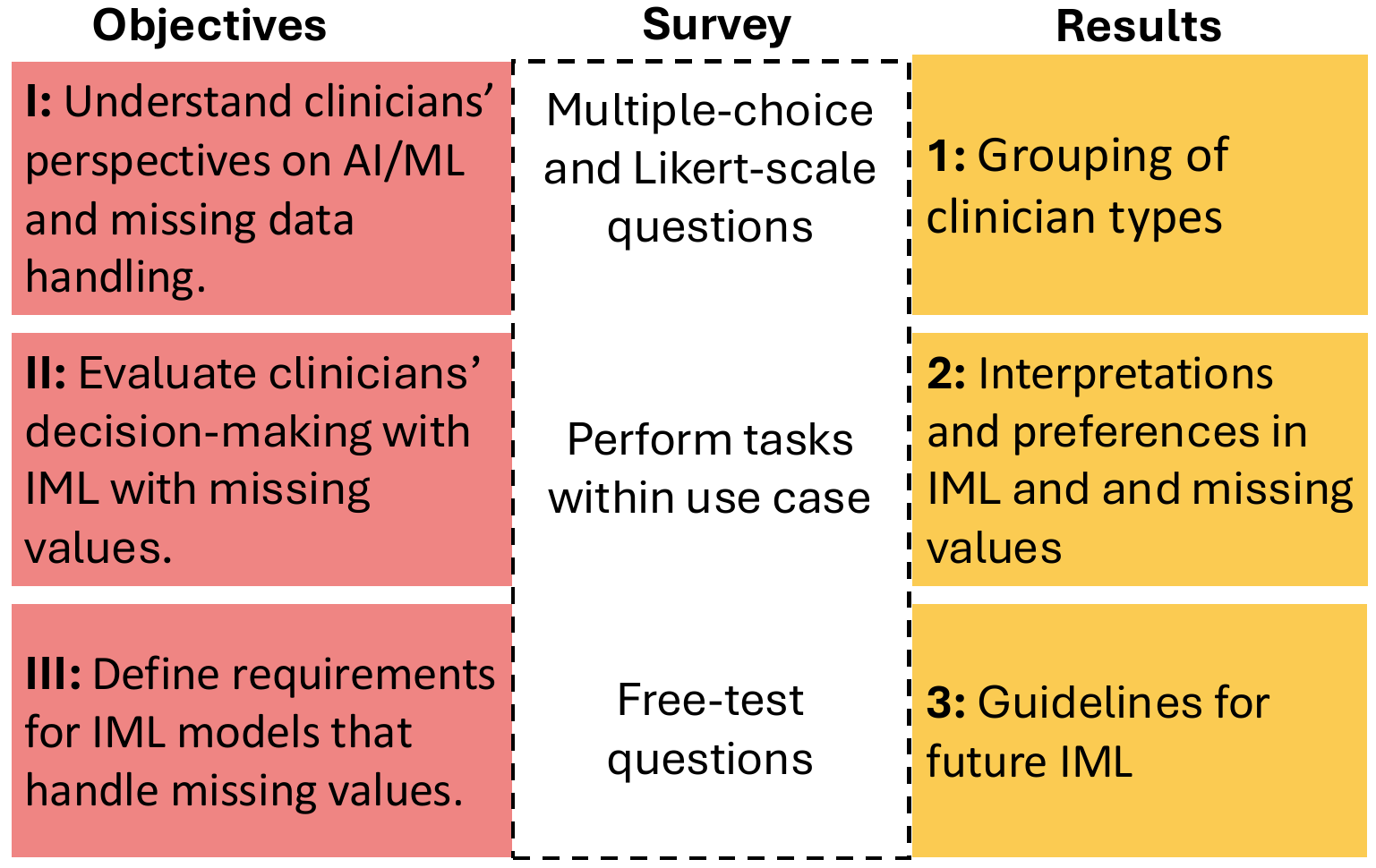}\vspace{-2em}}
\end{figure}

In this work, we survey clinicians to gain qualitative insights into their handling of missing values using IMLs, providing a real-world perspective on decision-making. Our evaluation is structured around three key objectives (shown in Figure~\ref{fig:objectives}). First, we aim to understand clinicians' attitudes toward artificial intelligence/machine learning (AI/ML) in healthcare and their current practices for clinical decision-making with missing values (Objective I). Next, we present a use case aimed at identifying common interpretations and usages of three different IML approaches and preferences for strategies to handle missing values (Objective II). Finally, we define requirements for future IML models that handle missing values at test time, ensuring they align with clinician preferences and can be seamlessly integrated into clinical workflows (Objective III).

Our survey focuses on a real-world scenario in handling missing values for interpretable clinical prediction. This complements and contrasts with the rich literature on quantitative and theoretical comparisons between, e.g., different imputation methods. We emphasize domain expertise through direct quotes and publish our full questionnaire, enabling collaborative research on IML and missing values to identify best practices across healthcare systems.

Our main contributions are as follows:
1) We clustered the survey responses of 55 survey participants to analyze clinicians' attitudes toward AI/ML and their practices for handling missing values, providing an overview of existing workflows. 2) We assessed clinicians' use of three different IML approaches in scenarios with missing values, analyzing their reasoning, preferences, and challenges in leveraging these models for decision-making. 3) Based on clinician feedback, we formulated guidelines for designing IML models that handle missing values natively, ensuring better alignment with clinical workflows and facilitating real-world adoption.

Our findings highlight a gap between missing value handling in machine learning research, which often employs overly simplistic or too complex imputation methods, and clinical decision-making, where clinicians prioritize observed data and medical intuition at the time of decision-making. Future approaches should focus on reducing reliance on imputation by adopting more interpretable strategies, with models that natively handle missing values. 

\section{Survey methodology}
\paragraph{Participant recruitment.}
We conducted an online survey using LimeSurvey~\footnote{\url{https://www.limesurvey.org}} in July–August 2024 to assess clinicians' understanding of IML models and handling of missing values in decision-making. The survey involved 214 clinicians from 29 trauma centers across France, all part of the TraumaBase~\footnote{\url{https://www.traumabase.eu/en_US}} network, which collects data on over 50,000 trauma cases annually. Responses were assumed to be largely independent due to the geographical spread of centers. To ensure clinical relevance, we focused on predicting hemorrhagic shock~\citep{dutton2010trauma}, a critical ICU scenario where missing information is common. In trauma ICUs, data is often incomplete due to lab results being unavailable before critical decisions are needed. Trauma ICU clinicians, accustomed to clinical scores that align with IML frameworks, are well-suited to evaluate IML models under test-time missingness. To boost response rates, we focused on a single use case and kept the survey within 20–25 minutes (details in Appendix~\ref{app:use_case}). Participants provided informed consent before participation.

\paragraph{Survey development.}
The survey was developed with the input of trauma physicians and machine learning researchers to ensure its relevance to both clinical and ML communities. To maintain linguistic accuracy and cultural relevance, it was provided in English and French by native speakers. The full survey in both languages is included in Appendix~\ref{app:survey_questions}. 

\paragraph{Survey design.}
The survey consisted of 41 questions covering demographics, multiple-choice questions, ranking tasks, Likert-scale statements, and open-ended questions, aligned with the survey’s three objectives (Figure~\ref{fig:objectives}). The \emph{first part} of the survey focuses on understanding clinicians’ perspectives on AI/ML models within healthcare and their current approaches to handling missing values during decision-making (Objective I). It includes multiple-choice and Likert-scale questions to capture detailed insights and measure attitudes consistently. Questions on current practices referenced trauma-related risk scores, such as the Red Flag~\citep{hamada2018development} or ABC score~\citep{nunez2009early}, ensuring familiarity and serving as a warm-up by linking familiar tools to the challenges of missing values.
The \emph{second part} of the survey examines a specific use case (Objective II), where clinicians predicted hemorrhagic shock for a patient with missing values using various IML models. It assesses how clinicians engaged with model outputs for decision-making, focusing on interpretability and handling incomplete data. Multiple-choice and ranking tasks uncovered preferences and priorities when interpreting IML models (details in the next paragraph).
Finally, the \emph{third part} comprises open-ended questions, allowing clinicians to share insights, contributing to guidelines for future IML research (Objective III).

\paragraph{IML and missing values in a use case.}
The second part of the survey presented a patient case with a missing value and predictions by three IML models (Figure~\ref{fig:IML_survey}). Participants were asked how they would handle the missing value in their decision-making process.
This allows us to investigate whether clinicians would rely on additional information, such as a subpopulation's average feature value, assign zero for the missing value, or apply their domain knowledge to estimate it. Additionally, we asked for their preferences in imputation strategies for different IML methods to better understand their approach to decision-making.

The survey asked questions about three well-established IML models: risk scores~\citep{ustun2019learning}, linear models~\citep{hastie2009elements}, and decision trees~\citep{breiman2017classification}. Since our primary interest was interpretability rather than predictive performance, we ensured all models had comparable accuracy~\citep{molnar2020interpretable, stiglic2020interpretability}. 

Handling missing values is challenging for these models, as they lack native support and often rely on imputation for predictions. Linear models require explicit handling of missing values since each feature contributes to the prediction based on its learned coefficient. If a feature is missing, its contribution is unknown unless imputed, meaning the model implicitly assumes the imputation strategy is correct. Rule-based risk scores require all features referenced in a rule to be available; otherwise, rule activation may be uncertain or impossible. Decision trees struggle with missing values due to their hierarchical structure, where each node depends on feature availability. If a feature is missing, the tree may follow a default branch (e.g., XGBoost~\cite{chen2016xgboost}) or use imputed values, both potentially introducing bias. While XGBoost learns the default direction during training, this approach lacks interpretability at test time. Alternatively, decision trees can use \textbf{surrogate splits}~\citep{feelders1999handling}, identifying alternative features that mimic the primary split, enabling predictions without explicit imputation.

We use methods feasible for handling test-time missingness: imputation and native approaches:
\begin{itemize}
    \item \textbf{Native methods}: These are model-inherent mechanisms for handling missing values, such as surrogate splits in decision trees, which rely on alternative features to make predictions~\citep{feelders1999handling}.
    \item \textbf{Implicit imputation}: Clinicians were presented with missing values without automated imputation and asked to estimate them using available features and their clinical judgment.
    \item \textbf{Explicit imputation}: These included simple statistical replacements, such as mean or zero imputation~\citep{rubin1976inference}.
    \item \textbf{Black-box methods}: More complex imputation techniques, such as Multiple Imputation by Chained Equations (MICE)~\citep{van2011mice}, were used to assess how opaque imputation models influence clinician trust and decision-making.
\end{itemize}

\begin{figure}[t]
    \floatconts
    {fig:IML_survey}
    {\caption{Clinicians were given a patient sample (top) and three interpretable ML models (middle: decision tree, logistic regression; bottom: risk score) predicting hemorrhagic shock. We assessed their interaction with the models with missing values.}%
    \vspace{-0.8em}}
    {%
        \centering
    \includegraphics[width=\columnwidth]{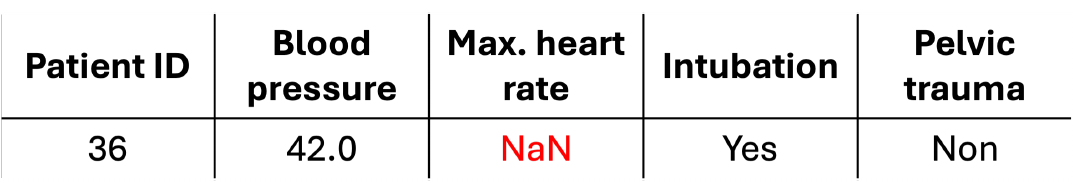}
        \vspace{-0.5em} 
        \begin{minipage}{0.45\columnwidth}
            \centering
            \includegraphics[width=\linewidth]{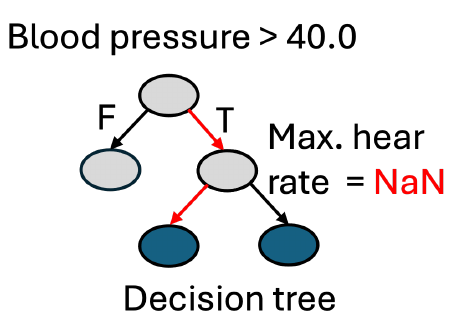}
        \end{minipage}
        \hfill
        \begin{minipage}{0.45\columnwidth}
            \centering
            \includegraphics[width=\linewidth]{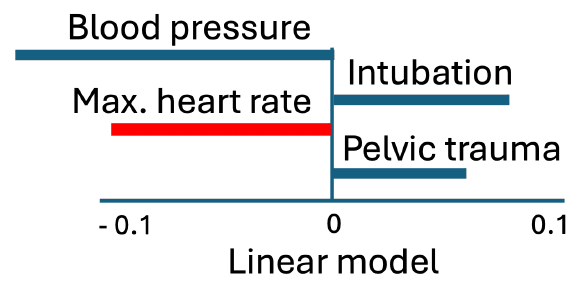}
        \end{minipage}

        \vspace{0.3em} 
        \includegraphics[width=0.5\columnwidth]{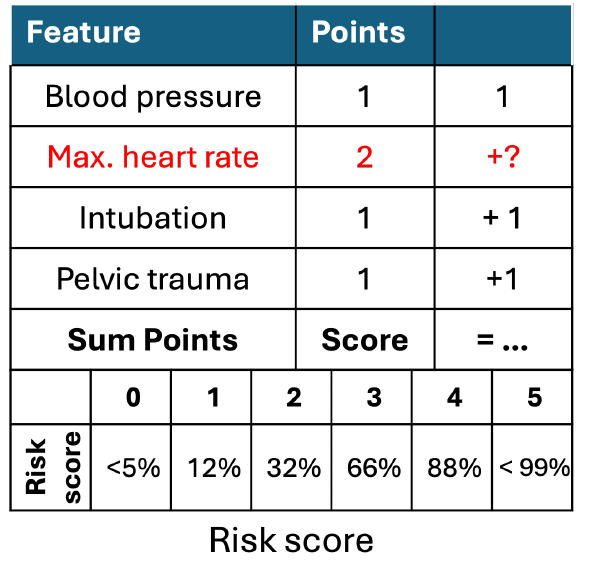}
        \vspace{-2em}
    }%
\end{figure}

\paragraph{Data analysis.}\label{subsec:data_analysis}

We analyzed multiple-choice responses using descriptive statistics. For questions with responses on the Likert scale or categorical responses, we applied Factor Analysis for Mixed Data (FAMD) in R~\citep{le2008factominer} to explore latent structures and identify response patterns. FAMD, the mixed data counterpart of Principal Component Analysis (PCA), accommodates both continuous and categorical variables, making it particularly well suited for questionnaire data~\citep{josse:hal-00811888}. Since our study focuses on missing values, FAMD provides a natural framework where handling missingness is integrated into the analysis. To prevent bias from discarding incomplete responses, we employed the Expectation-Maximization (EM) algorithm to estimate principal components while maximizing the observed likelihood. This approach ensures that missing value estimation aligns with imputation, where minimizing the weighted least squares criterion or maximizing the observed likelihood produces equivalent results. To further analyze clinician groupings (Finding~\ref{subsec:finding1}) based on AI/ML attitudes and incomplete data practices, we applied clustering to the FAMD results, allowing for a multivariate characterization of participants beyond isolated variables. Missing values in the survey answers, such as unanswered questions, were imputed using PCA for continuous variables and the most frequent category for categorical data. This process was implemented using the missMDA package~\citep{josse2016missmda}, which relies on iterative Singular Value Decomposition (SVD)~\citep{hastie2015matrix}, ensuring that imputed datasets preserve the same principal components as the original, maintaining the dataset’s underlying structure.
Free-text responses were analyzed for central tendency and variability. Figure~\ref{fig:survey_example} in the Appendix illustrates the full experimental setup.

\paragraph{Formal validation.}
We applied systematic methods to formally validate our qualitative survey and enhance the reliability, credibility, and accuracy of the data collection~\citep{rosellini2021developing}. First, we targeted domain experts to ensure the survey's relevance, clarity, and appropriateness. Before launching, we conducted a pilot study~\citep{shakir2022conducting} with five independent trauma clinicians, incorporating their feedback to refine the clarity of the question and the response options. In Section~\ref{sec:related}, we applied triangulation by comparing our findings with other data sources, such as quantitative studies, and literature. Finally, to ensure consistency in qualitative analysis, multiple researchers independently reviewed the coded responses.


\section{Best practices for IML with missing values at test time}
This section presents findings aligned with the three main objectives (Figure~\ref{fig:objectives}): clustering participants by their attitudes toward AI/ML in healthcare, summarizing clinician interactions with current IML models for handling test-time missingness, and outlining requirements for future IML research.

\subsection{Finding 1: Clustering based on AI/ML attitudes and missing values practices}~\label{subsec:finding1}
We received 55 survey responses, including 20 fully completed and 33 partially completed, all of which were analyzed, resulting in varying answer counts per question. The average participant age was 41.2 years, with a gender distribution of 22 females and 33 males. The majority of respondents were clinical doctors (83\%), primarily in Anesthesiology/Intensive Care (84\%), and 90\% held at least a medical degree. Participant demographics are provided in Table~\ref{tab:stats_participants}.

\begin{table*}[t] 
    \floatconts
        {tab:stats_participants}
        {\caption{Descriptive statistics of survey participants. Categorical variables have been simplified for brevity: The 'Other' category for current occupation included roles such as nurses, clinical studies technicians, hospital clinical trials coordinators, clinical manager assistants, and retired professionals. For medical specialization, it included fields like rehabilitation, pediatrics, surgery, anesthesiology, bioinformatics, pediatric orthopedic surgery, and clinical research associate.\vspace{-1.2em}}}
        {\begin{tabular}{lll}
        \toprule
        \textbf{Characteristic}  & \textbf{Total (N=55) } & \textbf{Complete answers (N=20)}\\
        \midrule
        Age in years, mean (SE) & 41.18 (1.13) &  39.35 (1.55) \\
        Female, n (\%) &  22 (40\%) & 7 (5\%) \\
        Male, n (\%) &  33 (60\%) & 13 (65\%) \\
        \midrule 
        Highest received educational degree, n (\%) &  &\\
        \quad University bachelor/master/postgraduate education & 9 (16,36\%) & 3 (15\%)\\
        \quad Medical doctoral degree& 37 (67,27\%) & 12 (60\%) \\
        \quad Associate Professor/Professor & 9 (16,36\%)  & 5 (25\%)\\
        \midrule 
        Current level of occupation, n (\%) & & \\
        \quad Clinical doctor & 46 (83,64\%) & 19 (95\%)\\
        \quad Intern/Trainee/Non-fully graduated clinician & 3 (5,45\%) & 0\\
        \quad Other & 6 (10,91\%) & 1 (5\%)\\
        \midrule
        Medical specialty, n (\%) & & \\
        \quad Anesthesiology $/$ Intensive Care& 46 (83.64\%) & 16 (80\%)\\
        \quad Emergency medicine & 5 (9.09\%)  & 2 (10\%) \\
        \quad Other  & 4 (7,27\%)  & 2 (10\%)\\ 
        \bottomrule
    \end{tabular}}
\end{table*}
We identified four distinct clusters among survey participants based on the FAMD results (see Section~\ref{subsec:data_analysis}). 
This provides a structured understanding of how clinicians’ attitudes toward AI/ML and their handling of missing values vary across different response groups (Figure~\ref{fig:attitude_for_AI_ML} and ~\ref{fig:cohort_cluster} in the Appendix). Cluster 1 includes clinicians (16\% of 55, 9 participants) with low understanding and use of risk scores, little familiarity with decision trees, and dissatisfaction with all three models when dealing with missing values. Despite this, they are comfortable using black-box models like MICE but prefer implicit imputation and surrogate splits. Cluster 2 consists of clinicians (22\%, 12 people) who show high levels of fear of AI, blind trust in its capabilities, and concerns about job loss, alongside perceiving current AI solutions as low-quality. They are familiar with decision trees and favor zero imputation or surrogate splits. Cluster 3 (36\%, 20 people) has significant missing data and may be excluded. Cluster 4 comprises clinicians (25\%, 14 people) who use AI daily, are highly familiar with all three IML models, and feel confident explaining these models with or without missing values. They favor implicit imputation over black-box models and surrogate splits.
These clusters highlight the diverse perspectives among clinicians, emphasizing the challenge of developing approaches that accommodate varying familiarity, trust, and confidence in handling missing data.

\paragraph{Current practices with missing values.}
Across all clusters, we identified various strategies to manage missing values in currently used clinical scores, e.g. RedFlag~\citep{dutton2010trauma}, primarily relying on clinical intuition or other available clinical information. Many mentioned using plausible values based on clinical context or the patient’s history to estimate missing values. Some prefer worst-case imputation to avoid underestimating risk, though one noted it may lead to overtraining and resource inefficiency. Others choose to partially calculate or omit scores if key data is missing, feeling that imputation could `skew the score concept'. One clinician also suggested using alternatives, like thromboelastography when prothrombin time is unavailable.

\subsection{Finding 2: Clinician preferences and interpretations from a use case with IML approaches and missing values} 

Next, we analyze clinicians' responses from the use case evaluation, focusing on their approach to handling missing patient data in IML. We summarize their interpretation patterns and preferences for imputation or native methods.

\paragraph{Confidence does not always align with accuracy when handling missing values.}
Out of 20 responses, 9 clinicians rated themselves as highly confident on a Likert scale in explaining risk scores, even with missing values, while 10 expressed the same level of confidence for decision trees. For logistic regression, however, confidence was lower and accompanied by uncertainty; only 6 participants were correct, with 4 out of 6 correct among those expressing confidence in its usefulness with missing values.

\paragraph{Low trust in zero and mean imputation.}\label{zero_mean_imputation}
When clinicians received the mean hemoglobin value for the dataset, almost all 23 respondents reported that this additional information neither influenced their decisions nor supported their decision-making in all three IMLs. The main reasons included the irrelevance to individual cases: ``I don't take it into account. Each patient is unique, and I don't see the relevance of giving him an average value.''
When presented with a prediction for a patient in which the missing value was imputed as zero, most clinicians indicated that they would ignore the feature in the risk score. Similarly, a machine learning model would also ignore the feature, effectively diminishing its influence on the prediction. While the representation of risk scores was still considered helpful by most, some clinicians preferred more accurate imputations for the linear model, with one remarking: ``If imputation is used during testing, it should be clearly indicated, ideally with the associated uncertainty. This transparency allows for comparison between predictions with and without the imputed value.'' 

\paragraph{Surrogate splits can be viable.}
For decision trees, using surrogate splits (the direction to take when the splitting feature is missing) was explored, but this method was less familiar to clinicians and caused some confusion. However, many clinicians noted that its usefulness depends on clearly indicating where the substitution occurs: “If the substitution is clearly indicated and can be taken into account by the clinician, yes, it could be a decision aid.” Despite this, some preferred other methods, one of them stating: ``Decision trees are easy to understand, but unstable and too sensitive to values. That's why we prefer Random Forest algorithms. It's difficult to rely on a decision tree, especially if there are missing values that have been imputed 
Rather than imputing [...], there are algorithms that estimate missing values in relation to other parameters.''

\paragraph{High trust in risk scores and decision trees despite missing values.} Lastly, the survey results show varying levels of trust among the 23 clinicians in using the three IML models for diagnosing hemorrhagic shock. Risk Scores were the most trusted, with 15 clinicians indicating they 'Almost Trust' the model. However, concerns about over-reliance on numbers rather than clinical judgment were noted, as one clinician remarked, ``I'm not dealing with numbers, but with a patient! I need a history, background treatment, a clinical history [...]'' Linear models faced the most skepticism, with 15 clinicians 'Not Sure' or 'Rather not' trusting them, due to unfamiliarity and doubts about handling missing values. Decision trees had mixed feedback—13 clinicians 'Fully or Almost Trust' them for their clarity, but some raised concerns about potential limitations. 

\paragraph{Comparison of imputation methods across IMLs.}

Clinicians were asked to rank commonly used machine learning imputation methods, comparing explicit approaches (e.g., zero, mean imputation) to implicit imputation, which relies on clinical intuition to infer missing values. The results indicate a slight preference for implicit imputation within risk score models, though explicit methods were also widely used. Notably, black-box imputation, such as MICE, was rarely chosen ($<$10\% of cases), despite its frequent use in the medical literature. This may stem from its lack of clinical interpretability and limited exposure to medical training. Figure~\ref{fig:imputation_IML} visually summarizes clinician selections across imputation methods and IML models. Surrogate splits and mean imputation were the most favored approaches, particularly in risk score and decision tree models, chosen in about 35-40\% of cases. We did not limit the selection to feasible combinations of IML and imputation methods. As a result, surrogate splits for risk scores and linear regression were selected, even though they lacked a sound methodological basis. Implicit imputation was more commonly selected for decision trees ($\sim$30\%), indicating its perceived alignment with tree-based models. Zero imputation, while moderately preferred (particularly in linear models, $\sim$25\%), was ranked lower than other explicit methods. These results highlight a strong clinician preference for transparent imputation techniques.

\begin{figure}[t]
\floatconts
  {fig:imputation_IML}
  {\vspace{-2.5em}\caption{Clinician preferences for imputation methods across different IML models. We normalize by dividing the number who chose a combination by the total, as the total votes for a model can vary.}}
  {\rotatebox{-90}{\includegraphics[width=0.65\columnwidth]{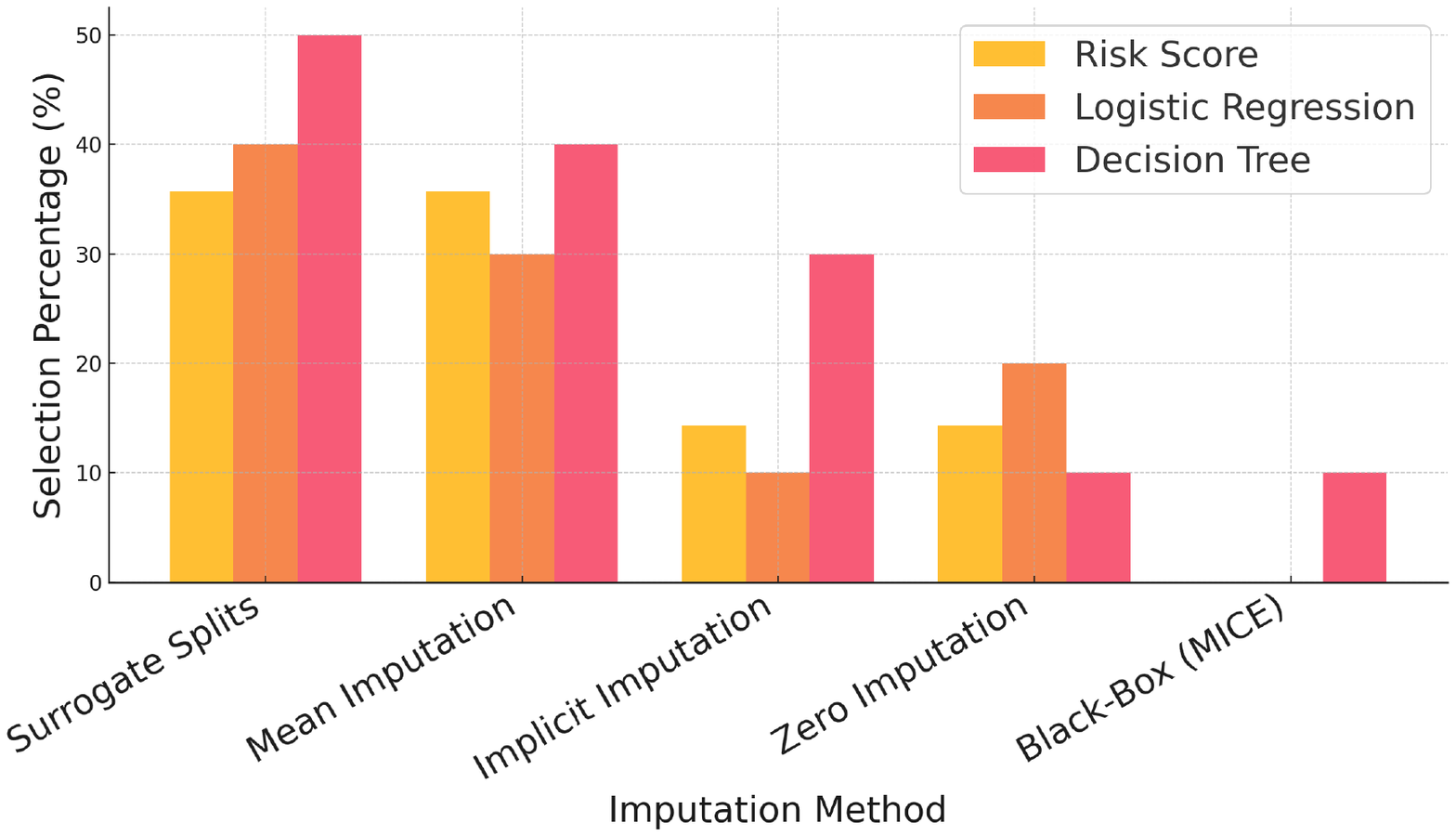}}}
\end{figure}

\paragraph{Clinically reasoned preferences for missing values at test time.} When choosing between implicit and explicit imputation methods, participants strongly preferred implicit imputation. This preference stems from the need for transparency, ease of understanding, and alignment with clinical reasoning, crucial in medical decision-making.
One participant emphasized these factors, stating, ``I prefer implicit imputation, as I'd find it hard to trust a black-box algorithm. Ease of understanding and communication are important, as well as from a 'medico-legal' point of view, to ultimately make a decision that involves my responsibility and the patient's life! [...] Even if the computer makes the decision, the responsibility [...] lies with me, and I must be able to understand and justify it.'' This highlights the importance of comprehensibility in high-stakes settings like healthcare.
Another respondent noted, ``I prefer implicit imputation because it seems simpler to reason with the values we have than to rely on a complex algorithm we don't understand.'' This reflects broader concerns among clinicians about the trustworthiness and transparency of black-box imputation.
The practical aspects of clinical work were also highlighted by a participant who remarked, ``Implicit imputation is more comprehensible and easier for clinicians to understand. Nevertheless, we need to see which technique is the best in terms of prediction, and I can't say what I prefer without knowing the relevance and PPV/VPn of each technique.'' This illustrates the balance clinicians must strike between understanding a method and ensuring its effectiveness.
Finally, the intuitive nature of implicit imputation was underscored by another participant, who stated, ``I prefer implicit imputation because, in my opinion, it's the closest thing to clinical reasoning: [...] it's the most intuitive and perhaps the safest approach, rather than an arbitrary rule.'' Some responses suggest sophisticated methods like MICE improve predictive performance by better preserving variable relationships than simpler methods like mean or zero imputation.

\subsection{Finding 3: Requirements for designing IML models for clinical integration}
Most respondents were open to using IML for diagnosis with missing values, though some were uncertain. A key concern was the need for \textit{clear guidelines} on handling missing values. Clinicians emphasized the importance of structured explanations, with one suggesting that models should present multiple imputed values for intuitive selection, while another stressed the need for ``a clear, simple, and rational explanation of how this missing data is treated, and its consequences on the test result.'' Without such transparency, trust in IML recommendations remains limited. Understanding and communicating \textit{uncertainty} also emerged as a critical factor. Clinicians highlighted the challenge of making decisions with missing values and emphasized the importance of incorporating decision thresholds. One respondent noted that they could make a decision despite missing values but would ``weight my decision with the patient's clinical data,'' underscoring the need for models to reflect confidence levels in their outputs. Beyond uncertainty, \textit{model reliability} remains a fundamental requirement. Clinicians stressed the importance of ensuring that both models and imputation methods produce stable and trustworthy results across different clinical scenarios. Without this assurance, the adoption of IML in practice would be challenging.
Equally important is the need for \textit{interpretability and accessibility}, particularly for non-technical users. Some clinicians highlighted the value of \textit{external validation} and visual aids to clarify how missing values impact model outputs, making results easier to understand and act upon.
Finally, \textit{clinical intuition must remain central to decision-making}. Clinicians emphasized that IML should enhance, rather than replace, human expertise. One respondent stressed that ``the human element and intuition based on experience and the CLINICAL EXAMINATION OF THE PATIENT!!!!'' should continue to guide clinical decisions, highlighting the need for models that complement expert judgment rather than override it.


\section{Related work}
\label{sec:related}
Our study aligns with previous findings that AI positively influences healthcare decision-making.~\citet{van2023intensive} show in their study, that 97\% of respondents were familiar with AI, and 86\% believed it could aid clinical work. However, concerns remain about confirmation bias, over-reliance on models, and the effort needed for tool interaction~\citep{wysocki2023assessing}. Our results suggest readiness for IML models, but successful implementation requires clinician participation in design to address needs like understanding missing data mechanisms and uncertainties~\citep{lage2019evaluation, lage2019human}. While \citet{janssen2010missing} argues that simple methods for handling missing data can lead to misleading results, others advocate for multiple imputation techniques as they enhance validity in clinical research \citep{austin2021missing}. In contrast, our expert evaluation suggests implicit imputation, based on clinical intuition, is generally preferred at test time, while black-box models are the least favored due to interpretability issues. Notably, ~\citet{read2024algorithms} highlights that senior critical care physicians have traditionally relied on autonomy and clinical intuition. While this approach enables adaptability in handling complex and unpredictable scenarios, it also contributes to unwarranted variations in care delivery and potential harm. 
Following~\citet{tonekaboni2019clinicians}, who emphasized evaluating clinical explainability in high-stakes settings, IML models must undergo rigorous validation to assess performance, interpretability, and legal implications~\citep{wiens2019no}. Clinicians can build trust in IML models by understanding how they work and through positive experiences where outputs align with their domain knowledge~\citep{kelly2019key}. Randomized controlled trials can validate models but are costly, time-consuming, and may lack generalizability~\citep{shortliffe2018clinical}. Our findings suggest that successful deployment requires robust and purpose-built models, a clear presentation of uncertainties~\citep{doshivelez2017rigorousscienceinterpretablemachine}, and training clinicians on the impact of missing values to integrate these tools into workflows~\citep{topol2019high, maddox2019questions}.

\section{Discussion and conclusion}\label{sec:discussion_conclustion}
Our survey evaluates how clinicians interact with predictive models, particularly in handling missing values. Our findings reveal a disconnect between missing data handling in ML and retrospective research, such as MICE imputation, and clinical decision-making, where clinicians rely more on observed data and medical intuition. While imputation methods like zero, mean, or black-box approaches may aid model training, they are less favored at test time. Worst-case imputation, though clinically cautious, can introduce statistical bias and misrepresent patient risk levels. 

To better align with clinical needs, future methods should either reduce reliance on imputation or adopt more transparent strategies. Native handling of missing values, as seen in tree-based models, can, for instance, be implemented using "Missingness Incorporated in Attributes" (MIA), which treats missing values as a distinct category~\cite{twala2008good,kapelner2015prediction, Josse2024}. This approach is similar to adding missingness indicators in linear models, although increasing the number of features can reduce interpretability. \citet{stempfle2024minty}, a generalized linear rule model minimizing reliance on missing values, avoids the need for imputation during training or testing, offering an intriguing direction for future research.

The diversity of clinician attitudes toward AI highlights the challenge of aligning models with clinical intuition and user preferences. Clinicians in Clusters 1 and 2, unfamiliar with IML, fearful of AI, or reliant on black-box models, would benefit from training on interpretable models, handling missing data, and AI's supportive role in practice. In contrast, Cluster 4, comprising highly engaged AI users, is well-suited for co-designing IMLs and refining missing value strategies. Capturing clinician intuition~\citep{read2024algorithms} by integrating preferred practices in a way similar to computer programming holds great potential, particularly in complex or evidence-limited scenarios. Without clinician involvement, the adoption of these tools is unlikely, underscoring the need for IML methods that enhance clinician interaction with predictive models, particularly for handling missing values.

This study lays the foundation for understanding the challenges in deploying clinical prediction models with missing values. However, it is subject to self-selection bias due to clinicians’ demanding workloads and limited availability. Future work could explore how these findings apply to other learning paradigms, like reinforcement learning, and different clinical settings, such as outpatient care,emphasising on the interaction between IMLs and clinicians. Expanding the study beyond the size of clinicians in France, we could also extend the study to include other healthcare systems for broader insights.



\bibliography{chil-sample}

\appendix

\section{Survey details}\label{apd:survey_details}

\paragraph{Survey setup}
LimeSurvey is an open-source online survey tool that allows users to create, manage, and analyze surveys. It provides an easy-to-access interface where participants can complete surveys via a web link. For your usage, participating clinics could access LimeSurvey to fill out surveys or collect data in a structured format. The tool offers customizable survey templates, various question types, and features like branching logic to tailor the survey experience, making it user-friendly and efficient for collecting data from clinical settings.

\begin{figure}[htbp]
\floatconts
  {fig:survey_example}
  {\caption{Experimental setup. Clinicians are shown a data entry of a patient with 5 features whereas one feature is missing in an interpretable machine-learning model along with the questions. After the answers are gathered qualitative and quantitative methods are used to analyze the results.}}
  {\includegraphics[width=0.9\linewidth]{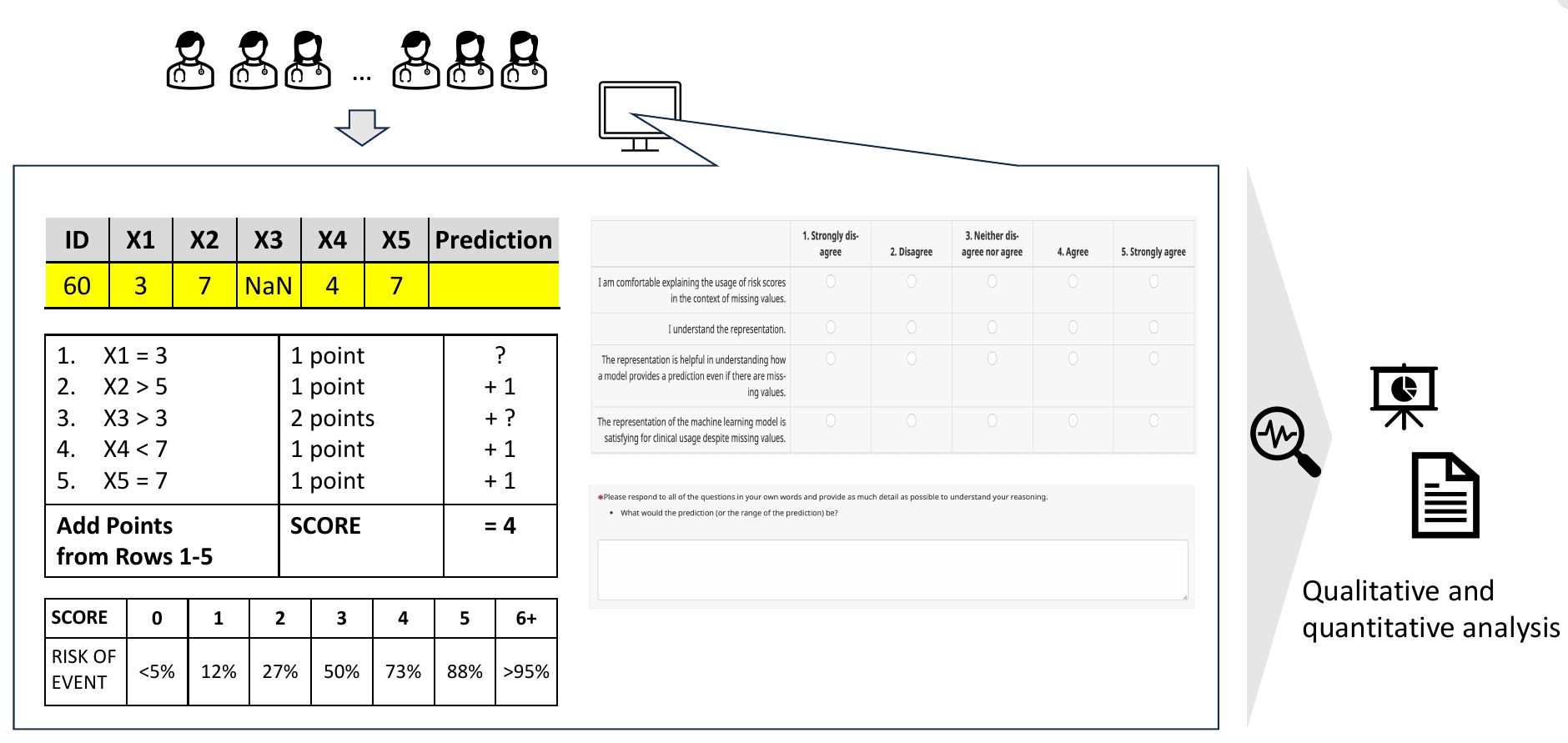}}
\end{figure}

\paragraph{Use Case of Hemorrhagic Shock}\label{app:use_case} 
Diagnosing hemorrhagic shock is often complex, as there is no universally accepted minimal set of criteria. Research, including a study by the Traumabase teams~\citep{james2022conundrum}, has highlighted the coexistence of multiple definitions of shock. In clinical practice, the diagnosis is based on a combination of factors indicating decreased cardiac output and hemorrhage. These include patient history (e.g., potential sources of bleeding), clinical examination (e.g., hypotension, tachycardia, visible bleeding), and paraclinical tests (e.g., active bleeding on CT scan or ultrasound findings). The survey use case was designed with input from trauma specialists, and we can clarify this further in the paper.

\paragraph{Missing values in FAMD}~\label{app:FAMD}
The primary goal of the Factor Analysis for Mixed Data (FAMD)~\citep{josse:hal-00811888} analysis and clustering is descriptive, allowing us to characterize clinicians from a multivariate perspective, rather than relying on just one or two variables to uncover patterns and group-specific insights. FAMD is the counterpart of PCA for mixed data, both continuous and categorical and is well-suited for describing results from questionnaires. Excluding incomplete responses could introduce bias if the missing data affects a specific subpopulation. Therefore, we compute the principal components using an EM algorithm, which maximizes the observed likelihood. In low-rank approximation methods like PCA or FAMD, estimation is equivalent to missing values and imputation. This means that minimizing the weighted least squares criterion or maximizing the observed likelihood yields the same results. The algorithms in missMDA~\citep{josse2016missmda} use an iterative SVD~\citep{hastie2015matrix} process: impute, compute SVD, update the imputation, and repeat. Thus, the imputation is "neutral" because the SVD on the imputed dataset produces the same results as on the dataset with missing values.

\section{Survey questions}\label{app:survey_questions}
The survey contained along with the demographic questions four domains. 
The domains are: Physicians’ Current Decision-Making Behavior in Diagnosing Hemorrhagic Shock in Trauma Patients (Current Decision-Making), Interpretation of ML models in the context of predicting with missing values at test time for a clinical use case (Interpretation of ML models and missing values), Preferences for using IML prediction model in daily workflows (Preferences in clinical care), Willingness to incorporate IML for prediction tasks in the context of missing values for diagnosing hemorrhagic shock and organ failure in clinical practice (Willingness)

In French: 
Les domaines sont les suivants: Comportement décisionnel actuel des médecins concernant le diagnostic du choc hémorragique chez les patients traumatisés (Opinion de l’Intelligence artificielle/Machine learning), Interprétation des modèles d'apprentissage automatique dans le contexte de la prédiction avec des valeurs manquantes au moment du test (Interprétation des modèles d'apprentissage automatique et des valeurs manquantes), Préférences pour l'utilisation du modèle de prédiction de l'IML dans les flux de travail quotidiens (Préférences de soins cliniques), Volonté d'incorporer l'IML pour les tâches de prédiction dans le contexte des valeurs manquantes pour le diagnostic du choc hémorragique et de la défaillance d'organe dans la pratique clinique (Volonté).

\begin{table*}[hbtp]
    \scriptsize
    \centering
    \floatconts{tab:rationales_english}
    {\caption{All questions in survey in the English language}}
    {\begin{tabular}{p{2.5cm}|p{14.5cm}}
        \toprule
        \textbf{Domain} & \multicolumn{1}{l|}{\textbf{Question}} \\
        \midrule
        Current Decision-Making Behavior & \makecell[l]{Q: Are you currently relying on decision support tools, such as risk scores, for diagnosing hemorrhagic shock in \\ trauma patients? If yes, which ones and how do they support you?}  \\
        & \makecell[l]{Q: Are you currently making use of any risk scores, e.g., RedFlag or ABC Score to support you in the diagnosis of  \\ hemorrhagic shock for trauma patients? If yes, please name the risk score and how you interpret the risk score when \\ applying it to patients with missing values.}  \\
        & \makecell[l]{Q: For which group of patients is the diagnosis of hemorrhagic shock the most difficult?}  \\
        & \makecell[l]{Q: Which factors/features are most important when diagnosing hemorrhagic shock in trauma patients?}  \\ 
        \midrule
        \makecell[l]{IML models and \\ missing values \\ at test time }& \makecell[l]{Q: Please respond to all of the questions in your own words and provide as much detail as possible to understand your \\ reasoning. What would your prediction (or the range of the prediction) be?}  \\
        & \makecell[l]{Q: The average hemoglobin value in this data set for females is 12.70 g/dl. Does that knowledge affect your prediction \\ assuming the patient is a woman?} \\
        & \makecell[l]{Q: Assuming the prediction made for the patient (no. 32450) was made by adding a zero for the missing value, which \\ resulted in a positive prediction indicating they will have hemorrhagic shock, is the representation of risk scores despite \\ the missing value helpful for your decision-making?} \\
        & \makecell[l]{Q: Do you trust the risk score? Motivate your answer briefly.}  \\
        & \makecell[l]{Q: Please respond to all of the questions in your own words and provide as much detail as possible to understand \\ your reasoning. What would the prediction (or the range of the prediction) be?}  \\
        & \makecell[l]{Q: The average hemoglobin value in this data set for females is 12.70 g/dl. Does that knowledge affect your \\ prediction assuming the patient is a woman?} \\
        & \makecell[l]{Q: Assuming that the coefficients were derived from a dataset completed using mean imputation, a \\ positive prediction was then made for the patient. Would you trust that the patient has a hemorrhagic shock?}  \\
        & \makecell[l]{Q: Do you trust the linear model? Motivate your answer briefly.}  \\
        & \makecell[l]{Q: Please respond to all of the questions in your own words and provide as much detail as possible to \\ understand your reasoning. What would the prediction (or the range of the prediction) be?}  \\
        & \makecell[l]{Q: The average hemoglobin value in this data set for females is 12.70 g/dl. Does that knowledge affect your prediction \\ assuming the patient is a woman?}  \\
        & \makecell[l]{Q: Assuming the prediction made for the patient (no. 32450) was made by replacing the missing \\ value using a surrogate split, a common approach in decision trees to handle missing values, which resulted \\ in a positive prediction indicating they will have hemorrhagic shock. In a surrogate split, an alternative value is \\ used to decide whether to go left or right in the tree. Is the representation of risk scores despite the missing \\ value helpful for your decision-making?}  \\
        & \makecell[l]{Q: Do you trust the decision tree? Motivate your answer briefly.}  \\
        \midrule
        Preferences in clinical care & \makecell[l]{Q: Missing values in decision trees and risk scores can be handled by implicit imputation, which means that a \\ value for the missing one can be imputed by considering the values of the other features. This might result in \\ individual values depending on the clinician. Rank which method—decision tree or risk score—you think implicit \\ imputation is more suitable for?}  \\
        & \makecell[l]{Q: Another alternative is to use so-called black box models that perform the imputation by using methods \\ that use complex, often non-transparent algorithms to fill in missing values in a dataset. Examples might be using\\  Multivariate Imputation by Chained Equations (MICE). For which interpretable model would you most \\ likely choose the black box imputation? Rank them according to your preference.}  \\
        & \makecell[l]{Q: Do you prefer implicit or explicit imputation and why? You can consider aspects such as ease of understanding, \\ usefulness in communicating your prediction, or intuitiveness.}  \\
        & \makecell[l]{Q: Which missingness handling methods do you like best for each IML? Mark 3 in total with an x and add 0 \\ in the other cells.}  \\
        \midrule
        Willingness & \makecell[l]{Q: What is lacking in interpretable ML models for effective clinical use when dealing with missing values?}  \\
        \bottomrule
    \end{tabular}}
\end{table*}

\begin{table*}[hbtp]
    \scriptsize
    \centering
    \floatconts{tab:rationales_french}
    {\caption{Toutes les questions de l'enquête sont en français.}}
    {\begin{tabular}{p{2.5cm}|p{14.5cm}}
        \toprule
        \textbf{Domaine} & \multicolumn{1}{l|}{\textbf{Question}} \\
        \midrule
        Opinion de l’IA/ML & \makecell[l]{Q: Utilisez vous actuellements des outils d’aide à la décision, tels que les scores de risque, pour diagnostiquer le choc \\ hémorragique chez les patients traumatisés? Si oui, lesquels et comment vous aident-ils?} \\
        & \makecell[l]{Q: Utilisez-vous actuellement des scores de risque, par exemple RedFlag ou ABC Score, pour vous aider à diagnostiquer \\ le choc hémorragique chez les patients traumatisés ? Dans l'affirmative, veuillez indiquer le score de risque et comment \\ vous l'interprétez lorsque vous l'appliquezà des patients dont les valeurs sont manquantes. }  \\
        & \makecell[l]{Q: Pour quel groupe de patients le diagnostic de choc hémorragique est-il le plus difficile?}  \\ 
        & \makecell[l]{Q: Quels facteurs/caractéristiques sont les plus importants lors du diagnostic du choc hémorragique chez \\ les patients traumatisés?}  \\ 
        \midrule
        Interprétation des modèles IML & \makecell[l]{Q: Veuillez répondre à toutes les questions avec vos propres mots et fournir autant de détails que possible pour\\  comprendre votre raisonnement. \\ Quelle serait votre prédiction (ou la plage de prédiction)?}  \\
        & \makecell[l]{Q: La valeur moyenne d’hémoglobine dans cet ensemble de données pour les femmes est de 12,70 g/dl. Cette \\ connaissance affecte-t-elle votre prédiction \\ en supposant que le patient est une femme?}  \\
        & \makecell[l]{Q: En supposant que la prédiction faite pour le patient (n° 32450) a été réalisée en remplaçant la valeur manquante \\ par un zéro, ce qui a \\ abouti à une prédiction positive indiquant qu'il y aura un choc hémorragique, la représentation des scores de risque \\malgré  la valeur manquante est-elle utile pour votre décision? Motivez brièvement votre réponse. }  \\
        & \makecell[l]{Q: Faites-vous confiance au score de risque? Motivez brièvement votre réponse.}  \\
        & \makecell[l]{Q: Veuillez répondre à toutes les questions avec vos propres mots et fournir autant de détails que possible pour \\ comprendre votre raisonnement. Quelle serait la prédiction (ou la plage de prédiction)?}  \\
        & \makecell[l]{Q: La valeur moyenne d’hémoglobine dans cet ensemble de données pour les femmes est de 12,70 g/dl. Cette \\ connaissance affecte-t-elle votre prédiction en supposant que le patient est une femme?}  \\
        & \makecell[l]{Q: En supposant que les coefficients étaient dérivés d'un jeu de données complété par imputation moyenne, une \\ prédiction positive a alors été \\ faite pour le patient. Seriez-vous convaincu que le patient présente un choc hémorragique?}  \\
        & \makecell[l]{Q: Faites-vous confiance au modèle linéaire?}  \\
        & \makecell[l]{Q: Veuillez répondre à toutes les questions avec vos propres mots et fournir autant de  détails que possible pour \\ comprendre votre raisonnement. Quelle serait la prédiction (ou la plage de prédiction)?}  \\
        & \makecell[l]{Q: La valeur moyenne d’hémoglobine dans cet ensemble de données pour les femmes est de 12,70 g/dl. Cette \\ connaissance affecte-t-elle votre prédiction \\ en supposant que le patient est une femme?} \\
        & \makecell[l]{Q: En supposant que la prédiction pour le patient (no. 32450) a été faite en utilisant une technique appelée division \\ de substitution pour traiter la valeur manquante, cette méthode est en effet couramment employée dans les arbres\\  de décision. Dans une division de substitution, une autre valeur est utilisée pour décider du chemin dans l'arbre \\(gauche ou droite). Imaginons qu'une division de substitution ait été effectuée et qu'elle ait conduit à une prédiction \\ positive indiquant un choc \\ hémorragique. La représentation des arbres décisionnels avec la division de substitution pour traiter la valeur \\ manquante vous aide-t-elle dans votre prise de décision?}  \\
        & \makecell[l]{Q: Faites-vous confiance à l’arbre de décision? Motivez brièvement votre réponse.}  \\
        \midrule
        Préférences de soins cliniques & \makecell[l]{Q: Les valeurs manquantes dans les arbres de décision et les scores de risque peuvent être traitées par imputation \\ implicite, ce qui signifie qu'une valeur pour la valeur manquante  peut être imputée en considérant les valeurs des\\ autres caractéristiques. Cela peut entraîner des valeurs individuelles en fonction  du clinicien. Classez quelle\\  méthode (arbre de décision ou score de risque) pour laquelle vous pensez que l’imputation implicite est la plus adaptée?}  \\
        & \makecell[l]{Q: Une autre alternative consiste à utiliser des modèles dits de boîte noire qui effectuent l’imputation en utilisant \\ algorithmes plus complexes, \\ souvent non transparents, pour combler les valeurs manquantes dans un ensemble de données. Des exemples pourraient \\ être l’utilisation de \\ l’imputation multivariée par équations chaînées (MICE). Pour quel modèle  interprétable choisiriez-vous le plus \\ probablement l’imputation par boîte noire? Classez-les selon vos préférences.}  \\
        & \makecell[l]{Q: Préférez-vous l’imputation implicite ou explicite et pourquoi? Pouvez-vous prendre en compte des aspects tels que \\ la facilité de compréhension, l'utilité de communiquer votre prédiction à un collègue ou l'intuitivité?}  \\
        & \makecell[l]{Q: Quelles méthodes de gestion des absences préférez-vous pour chaque modèle interprétable? Marquez vos 3 favoris \\ d'un x et ajoutez 0 dans les cellules.} \\
        \midrule
        Volonté & \makecell[l]{Q: Que manque-t-il aux modèles ML interprétables pour une utilisation clinique efficace lorsqu’il s’agit \\ de valeurs manquantes?} \\
        \bottomrule
    \end{tabular}}
\end{table*}

\section{Additional results}\label{apd:add_results}

This section presents further findings, including figures on the attitudes of different cohorts and the characteristics of identified clusters within these cohorts. 
\begin{figure*}[htbp]
    \floatconts
        {fig:attitude_for_AI_ML}
        {\caption{Cohort's attitude on AI/ML. We show the mode and standard deviation for eight statements exploring perspectives on AI/ML tools in healthcare, covering familiarity with these technologies, beliefs regarding their potential to replace physicians, expectations about their added value and support, and whether these tools adequately represent physicians' work to be useful. Missing values were not yet discussed.}}
        {\includegraphics[width=0.9\linewidth]{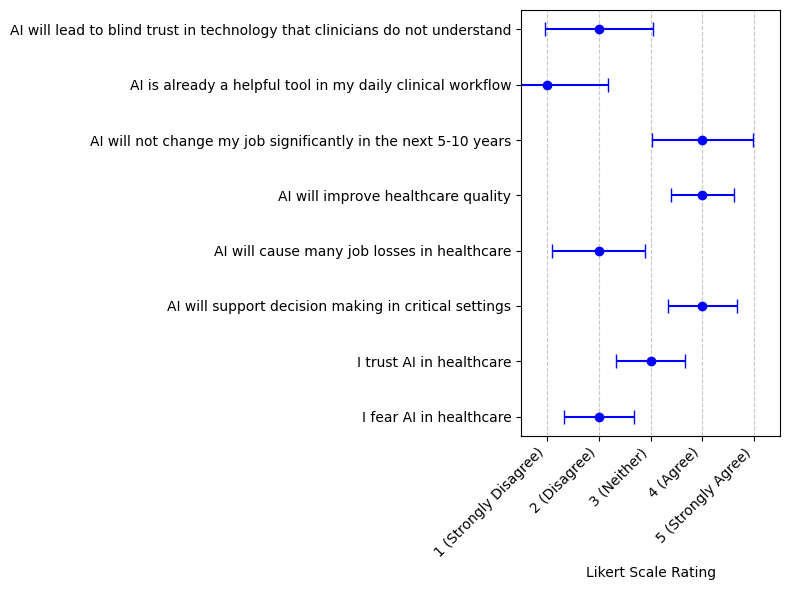}}
\end{figure*}

\begin{figure*}[htbp]
    \floatconts
      {fig:cohort_cluster}
      {\caption{The cohort is divided into four clusters, each reflecting different attitudes towards AI/ML, varying levels of familiarity with IML, both within and outside the context of missing values and imputation preferences along with general trust in IML. In the color coding, blue indicates that the mean for this cluster is significantly lower than the global mean, while red indicates that the mean is significantly higher. White or light shades of blue and red suggest no significant difference within the group regarding the variable. Demographical features were only added to describe the clusters.}}
      {\includegraphics[width=0.9\linewidth]{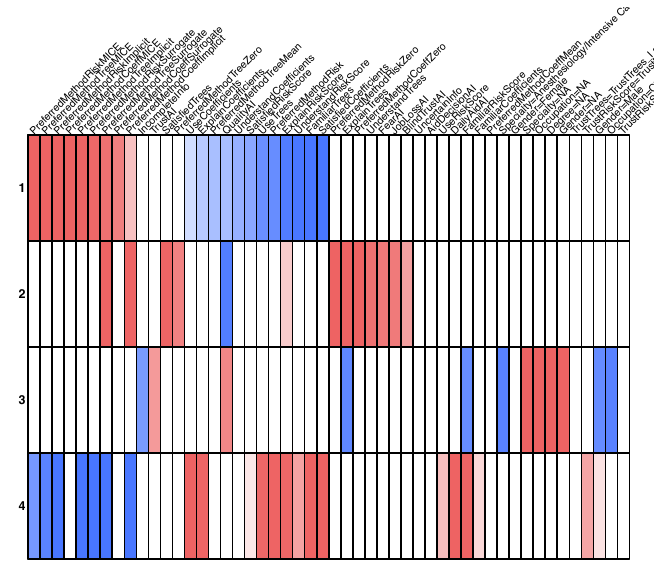}}
\end{figure*}

\end{document}